# Automated Literature Review Using NLP Techniques and LLM-Based Retrieval-Augmented Generation

Nurshat Fateh Ali
*Department of Computer Science and Engineering*
*Military Institute of Science and Technology*
Dhaka, Bangladesh
nurshatfateh@gmail.com

Md. Mahdi Mohtasim
*Department of Computer Science and Engineering*
*Military Institute of Science and Technology*
Dhaka, Bangladesh
mahdimohtasim@gmail.com

Shakil Mosharrof
*Department of Computer Science and Engineering*
*Military Institute of Science and Technology*
Dhaka, Bangladesh
shakilmrf8@gmail.com

T. Gopi Krishna
*Department of Computer Science and Engineering*
*Military Institute of Science and Technology*
Dhaka, Bangladesh
gopi.mistbd@gmail.com

*Abstract*—This research presents and compares multiple approaches to automate the generation of literature reviews using several Natural Language Processing (NLP) techniques and retrieval-augmented generation (RAG) with a Large Language Model (LLM). The ever-increasing number of research articles provides a huge challenge for manual literature review. It has resulted in an increased demand for automation. Developing a system capable of automatically generating the literature reviews from only the PDF files as input is the primary objective of this research work. The effectiveness of several Natural Language Processing (NLP) strategies, such as the frequency-based method (spaCy), the transformer model (Simple T5), and retrieval-augmented generation (RAG) with Large Language Model (GPT-3.5-turbo), is evaluated to meet the primary objective. The SciTLDR dataset is chosen for this research experiment and three distinct techniques are utilized to implement three different systems for auto-generating the literature reviews. The ROUGE scores are used for the evaluation of all three systems. Based on the evaluation, the Large Language Model GPT-3.5-turbo achieved the highest ROUGE-1 score, 0.364. The transformer model comes in second place and spaCy is at the last position. Finally, a graphical user interface is created for the best system based on the large language model.

## I. INTRODUCTION

Literature reviews have gained considerable importance for scholars. It provides researchers with a comprehensive overview of previous findings in a specific field and assists scholars in identifying gaps in past understandings. It helps to conduct future research and informs researchers of areas where they can provide significant input. However, conducting literature reviews can be incredibly cumbersome because there's so much to read. Due to the vast volume of research articles being released, reviewing all related studies and extracting relevant information can be a time-consuming, tedious, and error-prone task. Due to these difficulties, there has been an increasing interest in automating the process of literature reviews [1]. Automated systems can use natural language processing techniques and machine learning algorithms to analyze extensive amounts of text, extract relevant details, and create structured summaries [2].

The primary objective of this research is to develop a system that can automatically generate the literature review segment of a research paper by using only the PDF files of the related papers as input. Several Natural Language Processing techniques such as the Frequency-based approach, Transformer-based approach, and Large Language Model-based approach are implemented and compared to find the best procedure. The SciTLDR dataset [3] is selected for this research work. The first procedure uses the frequency-based approach. The library named spaCy [4] is utilized here. The second procedure uses the transformer-based model. The Simple T5 model is utilized here. The last procedure is based on using the Large Language Model. The GPT-3.5-TURBO-0125 model is utilized here. The evaluation and comparison are performed using ROUGE scores [5]. Then the best approach is identified and a Graphical User Interface-based tool is created.

Automating aspects of the literature review process allows academicians to save time and concentrate on the most pertinent articles for their research. It can also reduce the chance of errors or prejudice in the review process. The highlights of this article are:
- All three considered NLP approaches such as spaCy, T5, and GPT-3.5-TURBO-0125 model can produce satisfactory results in automating the literature review generation.
- The LLM-based model outperforms T5 and spaCy in generating literature reviews.

## II. LITERATURE REVIEW

A framework was proposed by Silva et al. [6] for automatically producing systematic literature reviews. They have focused on four technical steps: Searching, Screening, Mapping, and Synthesizing. In response to a specific inquiry, extensive searches are conducted to find as much relevant research as feasible, involving looking through reference lists, scouring internet databases, and reviewing published materials. Screening reduces the search scope by limiting the collection to only the papers pertinent to a particular review, aiming to highlight important findings and facts that could influence policy. Mapping is used to comprehend research activity in a particular area, involve stakeholders, and define priorities concerning the review emphasis. Synthesizing integrates data from numerous sources and provides an overview of the outcomes. The formulation of research questions, reporting phase, and peer review are some steps that are also discussed for the composition of systematic literature reviews.

Peer-reviewed publications are growing exponentially with the rapid development of science. Therefore, Yuan et al. [7] have explored the use of machine learning techniques, natural language generation, multi-document summarization, and multi-objective optimization for automating scientific reviewing. They have discussed the generation of comprehensive reviews and noted the limitations of constructive feedback compared to human-written reviews. The models used in this research are not yet fully capable of automating Literature Reviews and they require human reviewers.

A comprehensive analysis of existing tools for systematic literature reviews was done by Karakan et al. [8]. They have explored the potential for automation in various phases of the review process, highlighting the need for a holistic tool design to address researchers' challenges effectively. They have discussed two methodologies to accomplish their research: Rapid Review and Semi-Structured Interviews. Rapid Review emphasizes decision-making procedures for resolving issues, difficulties, and challenges that software engineers encounter in their daily work. Semi-structured interviews are used to explore researchers' experiences, challenges, strategies, strengths, weaknesses of Systematic Literature Review tools, and requirements for effective support in software engineering.

Jaspers et al. [9] focused on the use of machine learning techniques for automation of literature reviews and systematic reviews. They have outlined the pros and cons of different machine-learning techniques. The process of automating the literature review was elaborately discussed. The paper lacks practical validation across diverse domains and detailed insights.

A concise overview of automated literature reviews was presented by Tauchert et. al. [10] They have emphasized the potential for automation in various stages of the systematic review process. The paper discusses the importance of integrating computational techniques to streamline tasks such as searching, screening, extraction, and synthesis. It also acknowledges the need for further research to address challenges and enhance the effectiveness of automated approaches.

A brief overview on the topic of automatic literature review tools was given by Tsai et. al. [11] They discussed the existing research in the field, the challenges faced in conducting literature reviews manually, and the potential benefits of automating the process. The main focus of their contributions is the evaluation of Mistral LLM's effectiveness in the field of Academic Research.

The gaps in the intersection of systematic literature reviews (SLRs) and LLMs are discussed by Susnjak et. al. [12]. They also emphasized the need to address challenges in the synthesis phase of research and highlighted the potential of fine-tuning LLMs with datasets to enhance knowledge synthesis accuracy. The study aims to bridge this gap by proposing a Systematic Literature Review automation framework.

Most of the related works that have been discussed are mainly focused on discussing the potential and challenges of using NLP techniques and LLMs to automate the literature review process. None of them proposes a complete system pipeline where users can directly generate the literature review only using the PDF and DOI. In contrast, this article proposes and implements three unique end-to-end pipelines and procedures for a literature review automation system. This research endeavor has also resulted in the implementation of a UI tool where users can directly upload PDFs and get a literature review segment generated automatically without any additional effort. Moreover, this paper also includes a comparative analysis of different approaches such as the frequency-based approach, transformer-based approach, and rag-based approach using ROUGE scores which contributes towards finding the effectiveness of these approaches for this task.

## III. SYSTEM DESIGN

The research is carried out in four stages: 1. Defining research objectives. 2. Proposing multiple procedures for automated literature review generation. 3. Evaluating multiple procedures to find the best approach. 4. The final system development.

### A. Dataset Selection

The SciTLDR dataset from the Hugging Face is selected for this research work [13]. It contains the summarization of scientific documents. It is a dataset with 5,400 TLDRs derived from over 3,200 papers. It contains both author-written and expert-derived TLDRs of scientific documents. Curated research articles' abstract, introduction, and conclusion (AIC) or full text of the paper are given as "source" and the summaries of the corresponding articles are given as "target". Only these two attributes are utilized in all three proposed procedures. There is no training for the spaCy approach, but the dataset is utilized for testing purposes. The T5 model is trained using the SciTLDR dataset for the transformer-based approach and later evaluated on the test dataset. For the LLM-based approach, this dataset is used as the knowledge base for the model.

## B. The Procedure Utilizing the Frequency-Based Approach using spaCy

The first procedure utilizes the frequency-based approach by using spaCy. The first task is to build the model pipeline. The model pipeline takes text as input and converts the text into NLP tokens using the spaCy library. Then preprocessing step is done by removing stop words and punctuation. Afterward, the word frequency is calculated for each word which later helps to calculate individual sentence weights. This sentence weight represents the importance of that sentence. Then the top 10 percent of sentences are selected as the final output. The model is later evaluated using ROUGE scores to get an overview of the performance. The overview of the spaCy Model is given in Figure 1.

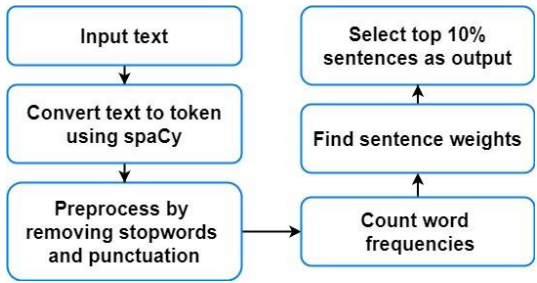

Figure 1: **Building spaCy Model**

The next step is to implement a system pipeline by using the spaCy model to generate a literature review segment automatically. The system takes the DOI and PDF files of multiple papers as input. It uses the Requests library to collect the paper titles and first author names from the DOI. Then it uses PYPDF2 and Regular Expression (RE) libraries to collect only the conclusion of each PDF. Then it uses the previously implemented spaCy model to get a summary of each paper. Later it performs post-processing and merges all summaries to produce a coherent literature review segment. The system pipeline overview of the spaCy Model is given in Figure 2.

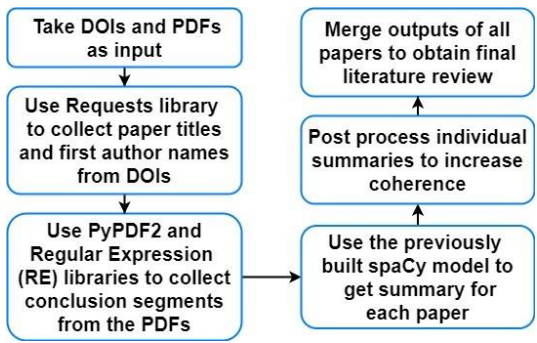

Figure 2: **Pipeline using spaCy**

## C. The Procedure Utilizing the Transformer-Based T5 Model

The second approach utilizes the transformer-based Simple T5 model. The first task is to train the model and prepare the model for the final pipeline. The SciTLDR dataset is collected to train the model. Then the dataset is prepared to use as the training data for the selected model. A task-specific prefix is added to summarize individual papers. Then the model is fine-tuned as per the requirements. Then the model is trained with the training data and the result is predicted. The result is the summarization of individual papers. Then the evaluation is performed using ROUGE scores and the model is saved for further utilization later in the system pipeline. The training overview of the Transformer Model is given in Figure 3.

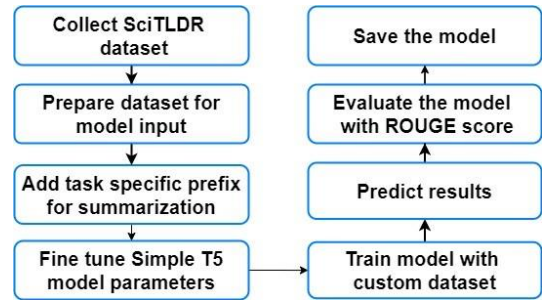

Figure 3: **Training of Transformer Model**

The next step is to implement a system pipeline by using the transformer-based model to generate a literature review segment automatically. The system takes the DOI and PDF of multiple papers as input. It uses the Requests library to collect the paper titles and first author names from DOIs. Then it uses PYPDF2 and Regular Expression (RE) libraries to collect each PDF's abstract, introduction, and conclusion. Then it merges 3 of these sections to get the final model input. Later it uses the previously trained and saved T5 model to get a summary of each paper. In the next step, it performs post-processing and merges all summaries to produce a coherent literature review segment. The system pipeline overview of the Transformer Model is given in Figure 4.

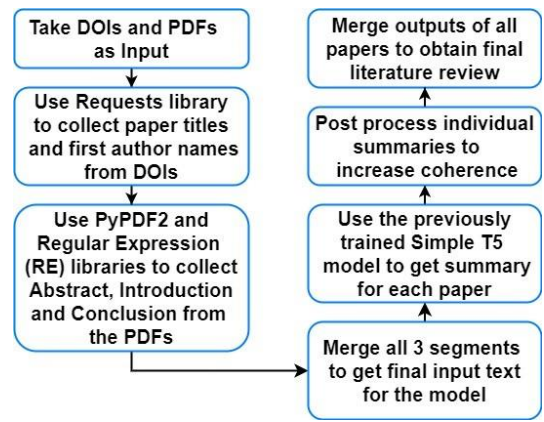

Figure 4: **Pipeline using Transformer Model**

*D. The Procedure Utilizing the Large Language Model: GPT-3.5-TURBO-0125*

The third procedure utilizes the RAG-based approach by using the Large Language Model: GPT-3.5-TURBO-0125. The first task is to create a custom OpenAI Assistant. Firstly, the SciTLDR dataset is collected, and then the GPT-3.5-TURBO-0125 model is selected for the OpenAI assistant. The retrieval is turned on and the dataset is added for the knowledge of the LLM. Now some prompt engineering is performed to produce the required output. Then the LLM results are evaluated using ROUGE SCORE. The overview of the creation of the OpenAI assistant is given in figure 5.

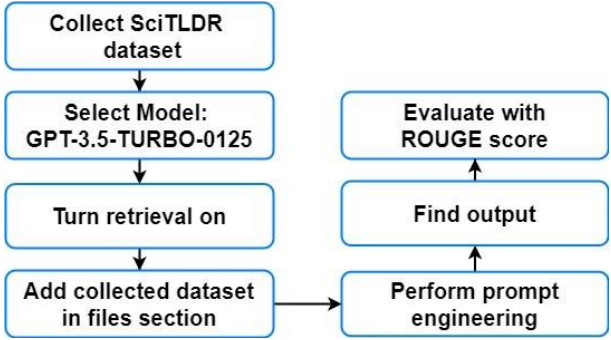

Figure 5: **Creation of Custom OpenAI Assistant**

The used prompt: "The user will give you a pdf file as input, similar to the "input" field of the given "data.json" file in your knowledge base. You have to produce a summarized "output" for the given pdf based on the file given to your knowledge. The output will be of max 80 words. Note: You must write in a way that can be considered a literature review of a new research paper. The user in the future might add more PDFs so try to make the literature review coherent and as per IEEE standards. Please mention the first author's name and paper title. Don't write like this "Literature Review of. . . "."

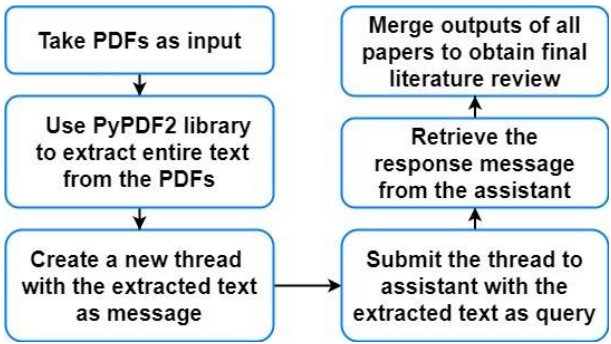

Figure 6: **Pipeline using LLM**

The next step is to implement a system pipeline by using the LLM to generate a literature review segment automatically. The system takes PDFs of multiple papers as input. It uses the PYPDF2 library to extract the entire text of each PDF. Then it creates a new thread with the extracted text as a message and submits the thread to the assistant with the extracted text as a query. Then the response from the assistant is retrieved and the outputs of each paper are merged for the final literature review segment. The system pipeline overview of the LLM is given in Figure 6.

*E. The Final System Tool*

The final system is implemented using the Large Language Model: GPT-3.5-TURBO-0125 as the backend. An aesthetic and simple user interface is created where the user can easily upload multiple research articles or PDF files. The user has to press the "Browse files" button and then select the files to upload. Then the system loads the research papers and within a few seconds, it produces the literature review segment automatically. It individually processes each paper and produces output. The loading screen and processing file numbers indicate the progress level and the number of processed papers. At the end of the literature review, the UI shows "Done" text to indicate the completion of the task. The user interface of the system is given in Figure 7

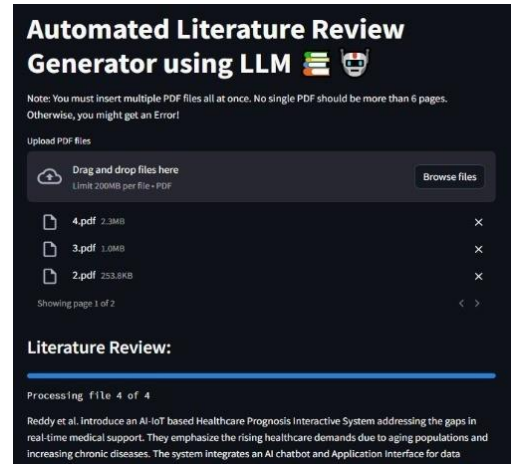

Figure 7: **The Preview of the System UI**

## IV. SYSTEM EVALUATION

The ROUGE scores are used for the evaluation in this research. The evaluation is done based on the test data of the selected dataset. ROUGE (Recall-Oriented Understudy for Gisting Evaluation) is a set of metrics used for evaluating the quality of machine-generated summaries by comparing them to reference summaries. The used ROUGE metrics are:

- ROUGE-N (precision, recall, and F1 score for n-gram overlaps),
- ROUGE-L (measuring longest common subsequence)
- ROUGE-Lsum (ROUGE-Longest for summary level evaluation)

*A. Evaluation of Frequency-Based spaCy*

The spaCy-based model was evaluated on the test data utilizing the ROUGE scores. The results are stated in Table I.

Table I: **ROUGE Scores for spaCy**

| ROUGE-1 | 0.257 |
|---|---|
| ROUGE-2 | 0.055 |
| ROUGE-L | 0.144 |
| ROUGE-L SUM | 0.146 |

*B. Evaluation of Transformer*

T5 The transformer-based model was evaluated on the test data utilizing the ROUGE scores. The results are stated in Table II.

Table II: **ROUGE Scores for T5**

| ROUGE-1 | 0.268 |
|---|---|
| ROUGE-2 | 0.115 |
| ROUGE-L | 0.204 |
| ROUGE-L SUM | 0.204 |

*C. Evaluation of Large Language Model: GPT-3.5-TURBO-0125*

The LLM-based model was evaluated on the test data utilizing the ROUGE scores. The results are stated in Table III.

Table III: **ROUGE Scores for LLM**

| ROUGE-1 | 0.364 |
|---|---|
| ROUGE-2 | 0.123 |
| ROUGE-L | 0.181 |
| ROUGE-L SUM | 0.182 |

*D. Comparison of Multiple Approaches*

The comparison between all three approaches is provided in Table IV.

Table IV: **Comparison of Rouge Scores**

|  | T5 | spaCy | GPT-3.5-TURBO-0125 |
|---|---|---|---|
| ROUGE-1 | 0.268 | 0.257 | 0.364 |
| ROUGE-2 | 0.115 | 0.055 | 0.123 |
| ROUGE-L | 0.204 | 0.144 | 0.181 |
| ROUGE-L SUM | 0.204 | 0.146 | 0.182 |

From the ROUGE scores, it is clear that the LLM-based model outperformed both T5 and spaCy. The Transformer-based model is in the second spot based on the ROUGE-1 score. The evaluation proves that large language models are superior to older NLP approaches.

*E. User Evaluation on UI*

The user evaluation was performed on the LLM-based UI system. The user chose 4 research articles as input. These were:
- AI-IoT based Healthcare Prognosis Interactive System
- Applying Internet of Things and Machine-Learning for Personalized Healthcare: Issues and Challenges
- Improving patients health-care using IoT
- IoT Based AI and its Implementations in Industries

The received output from the system: "The paper by Joshua Ernest Pedi Reddy et al. presents an AI-IoT based Healthcare Prognosis Interactive System (HPIS) to provide real-time medical diagnosis and support in the absence of healthcare workers. The system includes an AI-based chatbot and Application Interface to gather information, answer medical queries, assist users, and ensure medication adherence. With a focus on improving healthcare accessibility and quality, the system integrated smart pill dispensers for monitoring and enhancing medication adherence, offering great potential in revolutionizing healthcare services. Ahamed explores the application of IoT and Machine Learning in Personalized Healthcare (PH) to enhance disease management and intervention. IoT sensor devices and wearables collect patient data, which is analyzed using AI and ML techniques for disease prediction and patient self-management. Challenges include biased data collection, outdated training datasets, and privacy concerns. The integration of IoT and ML in PH faces issues such as data transmission reliability and biased training datasets, impacting diagnostic accuracy. Addressing these challenges is crucial for improving personalized healthcare systems. Khurana implements a Smart Healthcare System using IoT sensors to enhance patient care in hospitals. The system utilizes Ultrasonic and IR Proximity Sensors connected to an Arduino Uno for automated IV fluid level monitoring and patient alarm system. The literature review highlights the importance of automated Hospital Management Systems for efficient healthcare administration. The proposed system reduces manpower, costs, and human errors while improving patient care. Future research aims to enhance system scalability and functionality for widespread hospital use. Sherif El-Gendy explores the integration of IoT and AI in industries in the paper "IoT Based AI and its Implementations in Industries." The paper delves into Industry 4.0, IIoT, IAIoT, and IoRT, showcasing the impact on automation and robotics. It discusses IoT challenges, benefits of AI in data analysis, and presents case studies like oil field production optimization and smart robotics by companies like ABB and Boeing. The future of IoT/AI integration promises transformative advancements in various sectors."

## V. RESULT AND DISCUSSION

The study introduced three procedures for automated literature review generation. The research work also illustrates the performance comparison between various NLP approaches such as the frequency-based method (spaCy), transformer model (Simple T5), and retrieval-augmented generation (RAG) with LLM (GPT-3.5-turbo). All three procedures are implemented and the ROUGE-1, ROUGE-2, ROUGE-L, and ROUGE-Lsum scores are calculated based on the Test dataset. For all three approaches, the ROUGE-1 and ROUGE-2 scores are found above the acceptable mark.

From the evaluation, it is seen that the GPT-3.5-turbo model produced results with higher ROUGE-1 and ROUGE-2 scores than the SpaCy and T5. The overall ROUGE-1 score for the LLM is 0.364 while the score for T5 is 0.268 and spaCy is 0.257. It shows that the LLM-generated summaries have better unigram and bigram overlapping with human summaries. The transformer T5 is also an advanced model which comes in second place. The last position is occupied by the frequency-based spaCy model.

From the scores, it is clear that the most advanced models are LLMs which outperformed all other NLP techniques. But other approaches such as transformer models and frequency-based approaches are also capable of producing satisfactory ROUGE scores and a coherent literature review segment.

## VI. Conclusion and Future Scopes

The research focused on implementing and comparing various NLP techniques for automated literature review. All three implemented systems are successful in generating the coherent Literature Review segment of a research paper. The results of various Natural Language Processing techniques such as the Frequency-based approach, Transformer model, and Large Language Model are also successfully obtained and compared. Based on the comparisons, the LLM-based approach is proven to be the best-performing one based on ROUGE-N scores.

Thus, based on the LLM, a final system tool is also successfully developed where the user can upload multiple PDF files to automatically generate a coherent literature review segment.

Future work of this research work can be focused on enhancing the effectiveness and applicability of the developed system tool. More functionality can be added to the Graphical User Interface such as model options, output size, etc. More models such as Bert, Gemini, and LLaMA can be utilized to find better results.